\documentclass{neus2025}

\usepackage{amsmath}
\usepackage{booktabs}

\usepackage{booktabs}
\usepackage{graphicx}
\usepackage{times}
\usepackage{makecell}

\title[Gram-Space: Structure-Preserving NeSy Codebook Compression]
{Gram-Space: Structure-Preserving Codebook Compression for Memory-Efficient Neuro-Symbolic AI}

\author{%
 \Name{Weilun Wang} \Email{wwang346@ucr.edu}\\
 \addr University of California, Riverside
 \AND
 \Name{Wantong Li} \Email{wantong.li@ucr.edu}\\
 \addr University of California, Riverside%
}

\begin{document}

\maketitle
\begin{abstract}
Vector symbolic architectures (VSA) are widely used for reasoning in neuro-symbolic (NeSy) AI, yet high-dimensional codebooks often create severe memory bottlenecks that limit scalability and deployment. In this paper, we propose Gram-Space, a compression framework that applies Gram-Schmidt orthogonalization to represent codebook vectors in a compact orthonormal coordinate system. Gram-Space preserves the dot-product structure required by matrix-based VSA operators, which supports numerically equivalent execution of matrix similarity, probability vectorization, and attention score computations. We provide a correctness analysis showing that inner products are preserved under the orthonormal basis representation. 
Using modern GPU hardware, we benchmark the Gram-Space framework on standard neuro-symbolic reasoning datasets.
Experimental evaluations across state-of-the-art VSA models show that Gram-Space reduces model-level GPU memory usage by up to $15.75\times$ and improves inference latency by up to $3.62\times$. Profiling results further indicate that Gram-Space reduces allocation-heavy overhead in codebook-associated stages and improves hardware utilization for NeSy workloads.
\end{abstract}


\section{Introduction}
\label{sec:intro}

Neuro-symbolic (NeSy) AI is gaining significant momentum as large models encounter performance bottlenecks, particularly in handling edge cases and achieving high-precision reasoning~\cite{kautz2022third}. Though NeSy provides a practical alternative to purely neural approaches, the memory footprint of its inference pipelines constrains deployment on resource-limited hardware and reduces system-level efficiency~\cite{wan2024towards}. Within this domain, vector symbolic architecture (VSA) has demonstrated high model accuracy on complex reasoning tasks such as RAVEN and I-RAVEN~\cite{RAVEN,IRAVEN}. However, high-dimensional codebooks employed by VSA increase storage and memory traffic during similarity and probability-to-vector computations, which limits scalability across hardware platforms.
Traditional compression methods such as FFT-based approaches can reduce codebook storage, but they introduce unacceptable errors when applied to bipolar and binary codebooks common in VSA models~\cite{FFT}. Unlike existing hyperdimensional computing (HDC) compression methods that target classification workloads~\cite{morris2019comphd}, NeSy reasoning pipelines impose additional constraints including rule sparsity requirements and multi-stage dependencies between symbolic operators. Thus, these unique requirements motivate a compression approach that reduces memory overhead while remaining compatible with modern hardware.

To address these challenges, we propose Gram-Space, a compression framework that applies Gram-Schmidt orthogonalization to represent NeSy codebook vectors in a compact orthonormal coordinate system. While the underlying linear algebra is classical, its application to VSA-based NeSy models requires identifying which pipeline stages preserve correctness under the compressed representation and where reconstruction is necessary. The key observation is that codebook vectors occupy a subspace of dimension at most $M$, far smaller than the ambient dimension $D$, so matrix-based operations can be carried out entirely in this compact coordinate system without information loss. 
To our knowledge, Gram-Space is the first operator-aware codebook compression method for VSA-based NeSy reasoning pipelines that preserves matrix-space operations without information loss and requires no retraining.
Our main contributions include:
\begin{itemize}
\item We introduce a structure-preserving compression method for codebook-based NeSy models that represents codebook vectors in an orthonormal coordinate system. We provide a correctness analysis showing that inner-product structure is preserved exactly for matrix-space operators.
\item We propose an operator-space classification that distinguishes matrix-space operators from component-wise operators, which determines where compression applies and where reconstruction is required to preserve reasoning semantics.
\item We conduct operator-level profiling and roofline analysis that characterize how allocation and tensor materialization, rather than arithmetic computation, dominate memory pressure in codebook-associated stages.
\end{itemize}

\vspace{-10pt}
\section{Background and Related Work}
\label{sec:background}
\subsection{Computational Bottlenecks in VSA-based NeSy}

  \begin{figure}[t!]
      \centering      \includegraphics[width=\linewidth]{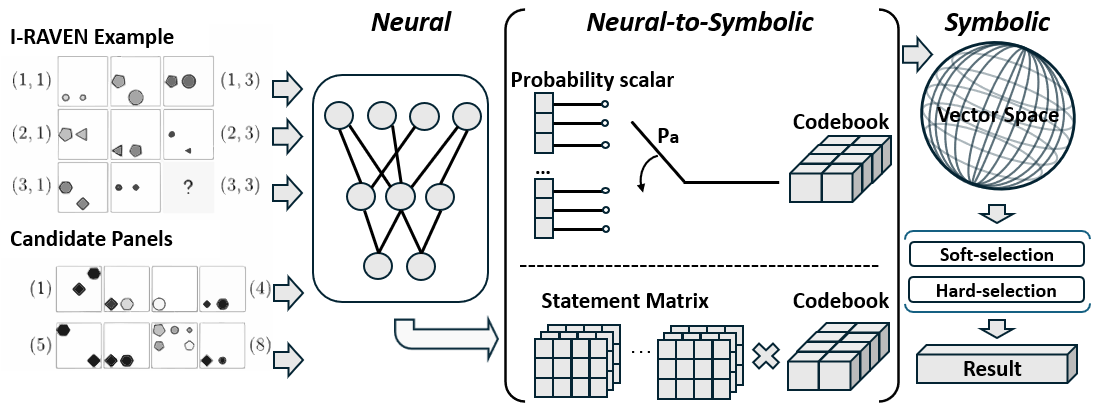}
      \caption{Overview of a codebook-based NeSy pipeline.}
      \label{fig:overview}
      \vspace{-10pt}
  \end{figure}

As illustrated in Fig.~\ref{fig:overview}, a VSA-based NeSy framework typically consists of three functional stages. First, a neural backbone performs representation learning and converts raw inputs into a unified form, such as probabilistic distributions or high-dimensional tensors. Second, the system aggregates these features through probabilistic blending or maps them to attribute-wise probabilities based on semantic similarity. Third, the model executes symbolic reasoning, where predefined rules operate on the processed representations to produce the final prediction. The codebook is the lexicon for the NeSy architecture and stores atomic high-dimensional vectors. Codebook vectors are initialized using a pseudorandom process (e.g. sampling from a Gaussian distribution or Bernoulli distribution). In high-dimensional spaces, two randomly generated vectors $\mathbf{x}$ and $\mathbf{y}$ are quasi-orthogonal with high probability. That is, their inner product is approximately zero: $\langle \mathbf{x}, \mathbf{y} \rangle \approx 0$.
To contextualize the problem, we summarize the representative NeSy models and components considered as case studies in this work.
NVSA is a hybrid framework that integrates a neural perception frontend with a VSA backend~\cite{NVSA}.
LearnVRF is a NeSy model that replaces static logic with learnable rule representations~\cite{learnVRF}. Unlike NVSA, it does not rely on rigid presets but instead optimizes rule matching through data-driven learning, which improves structural alignment and adaptability in reasoning tasks.
ARLC is a framework for rule exploration and flexible binding. It searches the rule space and emphasizes dynamic binding of attributes and relations rather than static matching~\cite{ARLC}.

  \begin{figure}[t!]
      \centering      
      \includegraphics[width=\linewidth]{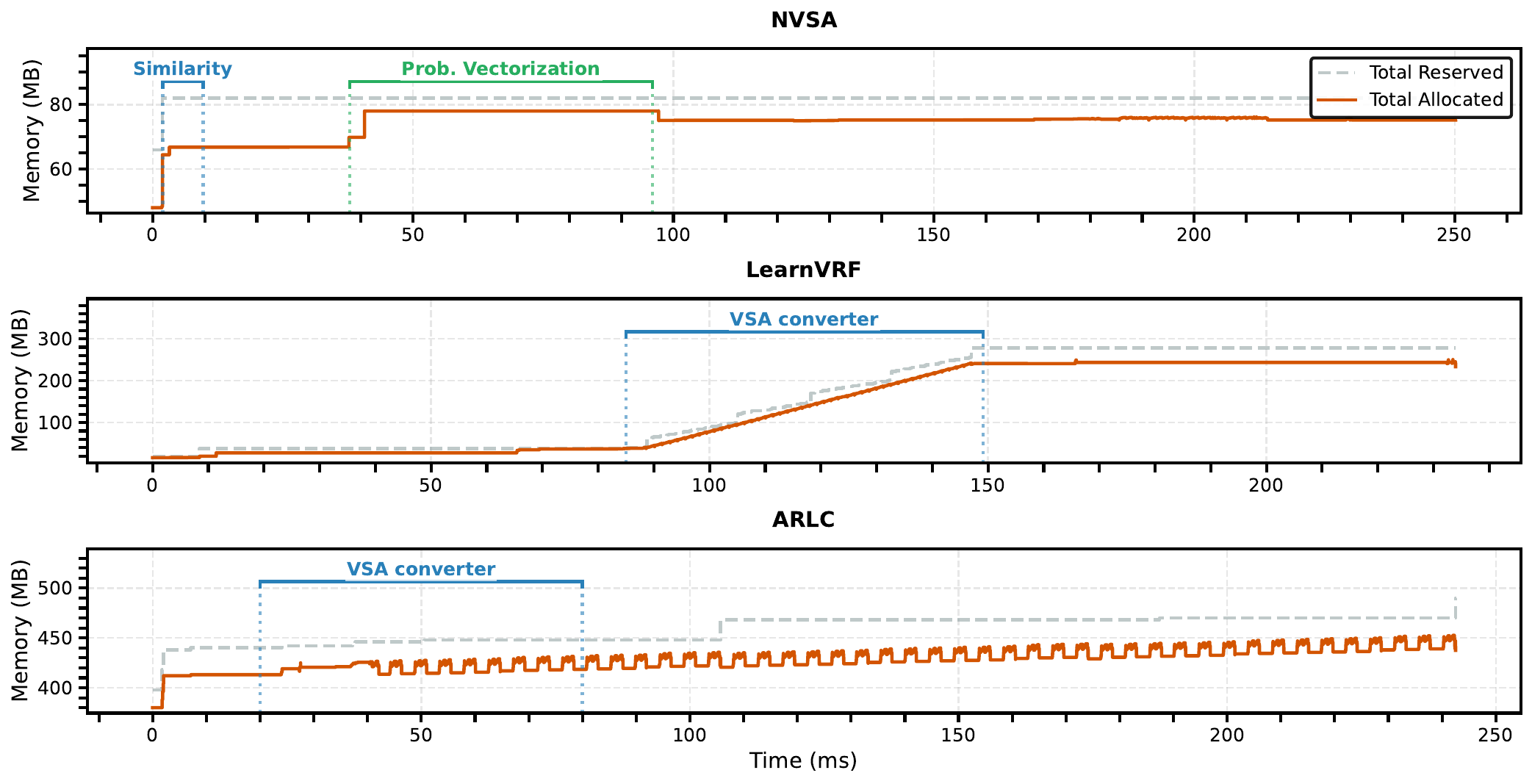}
      \vspace{-20pt}
      \caption{Memory traces for NVSA, LearnVRF, and ARLC during inference. Annotated regions indicate codebook-associated stages.}
      \label{fig:bottleneck_memory}
      \vspace{-30pt}
  \end{figure}

For these 3 models, we profile and record their memory traces during representative workload stages. The profiling is performed on an NVIDIA RTX 5070 GPU with 48 MB L2 cache. As illustrated in Fig.~\ref{fig:bottleneck_memory}, all 3 models exhibit a pronounced increase in memory consumption during a specific intermediate stage. The annotated regions correspond to stages that are strongly correlated with the codebook, and these regions show a clear spike in memory usage. This spike is substantially higher than the memory requirements of both preceding and subsequent phases, indicating a memory-intensive computational bottleneck. This high memory demand stems from the high-dimensional codebooks, which are required to maintain near-orthogonal properties. Even for datasets where atomic attributes are limited to tens of attribute types, the memory overhead remains substantial. This overhead increases hardware requirements and constrains scalability and deployment across platforms.

Existing NeSy codebook compression schemes typically fall into 3 categories: learning vector distributions via neural networks~\cite{onlineHD}, compressing information through stochastic methods~\cite{random}, or generating vectors based on single-task classification~\cite{singletask}. These approaches often impose substantial training cost, and they introduce information loss that increases with the compression ratio. Existing works on HDC compression and dimensionality reduction methods~\cite{morris2019comphd, dpqhd, sparsehd} have focused on classification tasks where accuracy degradation can be tolerated. In contrast, VSA-based reasoning pipelines require strict preservation of symbolic structure for rule detection and execution. As a result, existing methods are not well suited for settings that require high-precision symbolic reasoning, low training and inference overhead, and broad applicability across tasks.

\vspace{-10pt}
\subsection{Rule Detection and Execution in VSA}
\label{sec:workload_model}

Different stages in VSA rely on distinct domain-specific operations. The primary computational kernels in these stages are summarized here.
Probability vectorization transforms an inferred probability distribution into a representative vector by computing a weighted sum of attribute vectors from a codebook, where the weights correspond to attribute probabilities.
Scaled dot-product similarity compares vectors during inference through a scaled dot product between a normalized query vector $\hat{\mathbf{x}}$ and normalized codebook vectors $\hat{\mathbf{W}}_{j:}$, which maps directly to standard matrix multiplication.
Binding/unbinding is central to VSA rule execution. Its internal modulo-style computation places it in a different domain from matrix-based operations. In addition, binding and unbinding generate superposition states and practical VSA implementations often require high sparsity for reliable rule detection. 
Attention score selects among objects with different attributes~\cite{attention}. It computes weighted sums over vectors that may participate in multiple rules and supports selection among multiple candidate rule paths. In soft selection pipelines as described later, this stage separates vector-level aggregation from subsequent rule inference.

\vspace{-10pt}
\section{Proposed Gram-Space Compression}
\label{sec:prof}

\subsection{Gram-Space Compression for Codebook-based NeSy Models}
\begin{figure}[t!]
    \includegraphics[width=\linewidth]{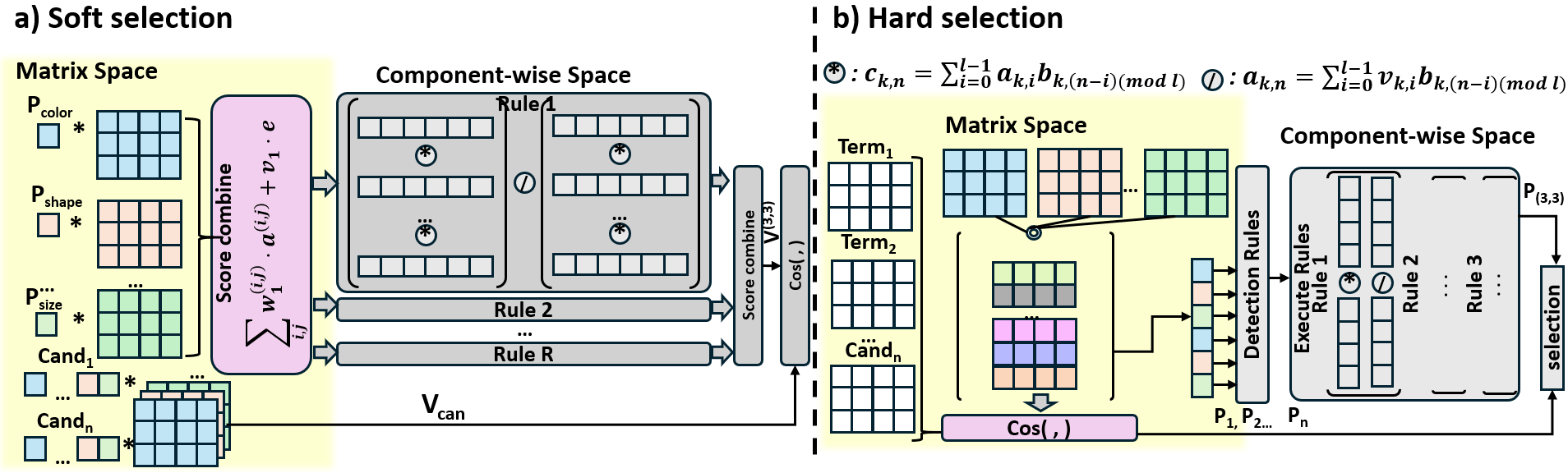}
    \vspace{-20pt}
    \caption{Operator-space classification and flow of Gram-Space.}
    \label{fig:space_classification}
    \vspace{-10pt}
\end{figure}

As illustrated in Fig.~\ref{fig:space_classification}, we categorize the primary computational operations in VSA-based NeSy into two paradigms: soft selection and hard selection. 
Soft selection functions as a probabilistic aggregation mechanism. It leverages probability distributions, typically generated by neural networks, to synthesize vectors via weighted superposition. Next, it computes inter-vector correlations through attention scores and executes flexible rule generation via binding and unbinding operations.
Conversely, hard selection adopts a similarity-based retrieval approach. In this paradigm, neural networks generate target vectors that are queried against a frozen, high-dimensional codebook to derive attribute probabilities. To ensure a comprehensive representation of all features, the total number of codebook entries, denoted by $M$, is defined by the cumulative aligned length of all atomic attributes. In the case of the RAVEN and I-RAVEN datasets, which involve 4 primary attributes with each aligned to a size of 10, the codebook comprises $M=40$ distinct atomic vectors. Furthermore,  $D$ is a large hyperparameter to ensure approximate orthogonality between representations. A parameter ${k}$ is then used to facilitate compression and mitigate dimensionality constraints when the available space is smaller than the class count. Finally, downstream operations such as superposition are performed to infer attribute probabilities, estimate rule likelihoods under predefined logic, and select the final output component.

Let the codebook be represented by a matrix $\mathbf{W}\in\mathbb{R}^{D\times M}$, where each column is an atomic codebook vector of dimension $D$, and $M$ is the number of codebook entries. Gram-Space constructs an orthonormal basis $\mathbf{U}\in\mathbb{R}^{D\times k}$ that spans the column space of $\mathbf{W}$, where $k$ is the algebraic rank of the codebook matrix. We compute the Gram-Space codebook projection $\mathbf{T}$ as
\vspace{-5pt}
\begin{equation}
\mathbf{T} = \mathbf{U}^{\top}\mathbf{W} \in \mathbb{R}^{k\times M},
\vspace{-5pt}
\end{equation}
and represent each codebook vector $\mathbf{x}$ by its coefficient vector $\mathbf{t}=\mathbf{U}^{\top}\mathbf{x}$. To reduce the computational complexity of codebook recovery, each atomic vector undergoes normalization.
We further categorize NeSy operators to belong to either the matrix space or the component-wise space. In matrix space, matrix-based computations operate on Gram-loc, which is the coefficient form obtained by projecting vectors into the Gram-Space basis. 
We set $\mathbf{U}$ as the Gram-Space basis, and the coefficient representation $\mathbf{t}$ or the corresponding column in $\mathbf{T}$ as the Gram-location (Gram-loc) representation. In matrix space stages, the model operates on Gram-loc for similarity measurements, transformations between probability vectors, and attention score calculations. Before the computation enters component-wise symbolic operators such as binding and unbinding, the representation is restored to the original domain by
$\mathbf{x} = \mathbf{U}\mathbf{t}$.
The subspace dimension $k$ is configurable: we set $k=M$ for maximal compression, or $k > M$ to preserve representation sparsity. In the latter case, the orthonormal basis is augmented via zero-padding and linear transformations to redistribute the signal uniformly across the expanded dimensions. 
When $D \geq M$, the codebook has full column rank and Gram-Space applies directly. When $D < M$, the codebook is rank-deficient, so we partition the $M$ entries into subsets of size at most $D$ and construct a separate orthonormal basis for each subset. After reconstruction, the subsets are concatenated to recover the full codebook layout.
This representation provides a direct memory model, where the original codebook storage scales as $O(MD)$. Under Gram-Space, the stored representation consists of the basis and coefficients, with storage $O(Dk+Mk)$. 
Therefore, Gram-Space yields net memory reduction when $k \ll D$. In VSA-based NeSy models, this condition holds by a wide margin: the number of codebook entries $M$ (and hence $k \leq M$) is determined by the attribute size, which is typically on the order of tens, while $D$ can scale to thousands. For the RAVEN and I-RAVEN benchmarks used in our experiments, $M = 40$ and the smallest $D$ is 256, yielding a theoretical ratio $D/k \geq 6.4$. As $D$ increases, the compression benefit grows proportionally.

\subsection{Correctness Analysis and Proof}
\label{sec:correctness}

We establish that the Gram-Space representation preserves the mathematical structure required by matrix-space operators in the NeSy pipeline. Gram-Space targets numerical equivalence for the matrix-based operations that dominate codebook-associated stages. Unlike stochastic compression schemes that can introduce noise that accumulates across pipeline stages~\cite{GHRR, FFT-based, onlineHD}, the orthonormal basis representation used in Gram-Space preserves inner-product structure in exact arithmetic. We analyze the invariance property for matrix space operators below.

Let $\mathbf{U}\in\mathbb{R}^{D\times k}$ be an orthonormal basis for the subspace used by the codebook, with $\mathbf{U}^{\top}\mathbf{U}=\mathbf{I}_k$. In our construction, codebook vectors and all intermediates produced by matrix space operators lie in $\mathrm{span}(\mathbf{U})$ by construction, since these intermediates are linear combinations of codebook columns. For any vectors $\mathbf{A},\mathbf{B}$ that lie in this subspace, inner products are preserved when $\mathbf{A},\mathbf{B}\in\mathrm{span}(\mathbf{U})$ under the basis representation:
\vspace{-5pt}
\begin{equation}
\langle \mathbf{A},\mathbf{B}\rangle = \langle \mathbf{U}^{\top}\mathbf{A}, \mathbf{U}^{\top}\mathbf{B}\rangle
\vspace{-5pt}
\end{equation}
This is a standard property of orthonormal basis transforms. As a result, dot-product similarity and cosine-style comparisons implemented through dot products are invariant under the Gram-Space representation. The same property extends to linear combinations of codebook vectors, and to attention score computations that use matrix multiplication. This yields the invariance relation:
\vspace{-5pt}
\begin{equation}
\text{Sim}(\mathbf{A}, \mathbf{B}) = \text{Sim}(\mathbf{U}^{\top} \mathbf{A}, \mathbf{U}^{\top} \mathbf{B})
\vspace{-5pt}
\end{equation}
for similarity functions $\text{Sim}(\cdot,\cdot)$ that depend on inner products and norms. Operators that require component-wise structure, such as softmax and binding/unbinding, are not executed in Gram-Space. For these operators, the model restores the representation to the original domain before execution.

We define the rank $k$ as the algebraic rank of the codebook matrix (or the dimension of the subspace spanned by the codebook vectors). Unlike low-rank approximations that discard tail components, Gram-Space constructs a complete orthonormal basis $\mathbf{U} \in \mathbb{R}^{D\times k}$ that fully spans the codebook. For any codebook vector or linear combination, the decomposition is exact:
\vspace{-5pt}
 \begin{equation}\mathbf{b}_{new} = \sum_{i=1}^{k} \langle \mathbf{b}_{new}, \mathbf{q}_i \rangle \mathbf{q}_i
\label{fun:gram}
\vspace{-5pt}
\end{equation}
Since $\mathbf{b}_{new} \in \mathrm{span}(\mathbf{U})$ by definition, the residual term is theoretically zero. Thus, the reconstruction does not discard any information in exact arithmetic: $\mathbf{x} = \mathbf{U}\mathbf{U}^{\top}\mathbf{x}$. The compression in Gram-Space arises strictly from exploiting the redundancy ($k \leq M$) inherent in the codebook geometry, rather than from numerical truncation.
In finite-precision arithmetic, the orthonormal basis is computed via Householder QR factorization, which is backward stable and avoids the loss of orthogonality associated with classical Gram-Schmidt.
Overall, Gram-Space projects codebook vectors into a dense coefficient space for matrix space operators and restores vectors to the original domain for component-wise symbolic operators. This design reduces memory traffic in matrix-heavy stages while preserving the dot-product structure required for similarity and probability-to-vector transformations.

\begin{figure}[t!]
    \includegraphics[width=\linewidth]{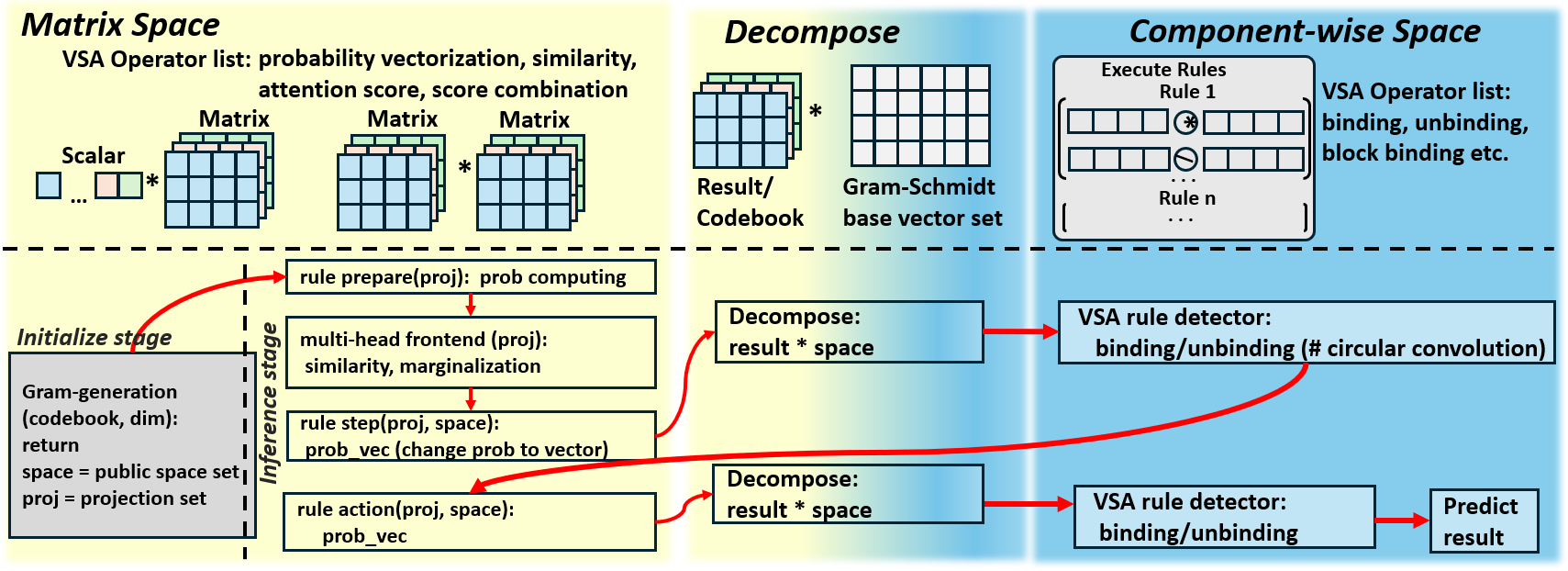}
    \caption{Detailed computing flow of Gram-Space to reduce hardware memory usage.}
    \label{fig:ComputingFlow}
    \vspace{-10pt}
\end{figure}

\subsection{Application of Gram-Space to Reduce Memory Traffic}
Fig.~\ref{fig:ComputingFlow} uses NVSA as an example to illustrate a practical Gram-Space flow. 
When the computation enters component-wise space, the model restores the required vectors to the original domain by multiplying the Gram-Space basis by the coefficients.
In practice, the rule stage often uses irregular data structures, and reconstruction cannot be expressed as a single global rule across all models. Thus, we place reconstruction at model-specific transition points where outputs of matrix space operators become inputs to component-wise operators. These transitions occur immediately before binding and unbinding, and at stages that require sparse representations for rule execution.
During initialization, the Gram-Space basis $\mathbf{U}$ and Gram-loc coefficients $\mathbf{T}$ are generated by Gram-Schmidt orthogonalization of the codebook. Prior to basis construction, the codebook is iteratively re-initialized to become fully orthogonal. During inference, multi-head similarity and probability computation operate directly on Gram-loc. Before rule execution, the model restores the vectors required by the rule detector by applying the reconstruction $\mathbf{x}=\mathbf{U}\mathbf{t}$. This procedure constitutes the decompose operation. Although the Gram-Space preserves perfect orthogonality, the subsequent operations based on circular convolution fundamentally rely on high sparsity to accomplish attribute superposition. Thus, while retaining perfect orthogonality does not compromise inference accuracy, restoring the sparse state is mandatory if vector-based inference is to be employed. Conversely, this reconstruction is redundant for models relying solely on similarity or probability outputs, allowing the entire inference to proceed within the coefficient space. Crucially, even when reconstruction is required, it is applied per-batch only to the active intermediate vectors. This implies that Gram-Space effectively reduces memory traffic by maintaining a global $\mathbf{U}$ and propagating only the compressed coefficients $\mathbf{T}$, regardless of the computational density in the matrix space.

\vspace{-10pt}
\section{Methodology and Results}
\label{sec:results}
\subsection{Experimental Methodology}

We evaluate the compression performance of Gram-Space and its implications for hardware memory systems using 3 representative models: NVSA, ARLC, and LearnVRF. Specifically, we employ PyTorch Profiler and CUDA tools on an RTX 5070 GPU to capture memory usage, task traces, and event latencies. This enables a comparative analysis of memory behavior pre- and post-compression. Furthermore, we derive operator density and atomic operations from the profiling traces, combining these statistics with memory metrics to estimate operational intensity and construct a roofline model for the RTX 5070. Since the neural backbones of the examined NeSy models remain unmodified, our profiling exclusively targets the VSA and symbolic rule execution components. Matrix-space operators utilize the original precision, while reconstruction aligns with the precision of the corresponding downstream stage. All results are reported under the default PyTorch execution mode. To eliminate interference from system-level overheads (e.g. excessive tensor slicing and Python-based CPU-to-GPU data transfer), we additionally conduct isolated benchmarks on general-purpose computational tasks using only the codebook. 
Prior to data collection, we execute multiple inference loops to warm up the GPU, thereby minimizing variability caused by hardware states like frequency scaling. To ensure timing accuracy, we measure latency and memory usage separately. CUDA Events are used to measure execution delay, while the Profiler is called in a separate run to generate memory reports.
\begin{figure}[t!]
    \includegraphics[width=\linewidth]{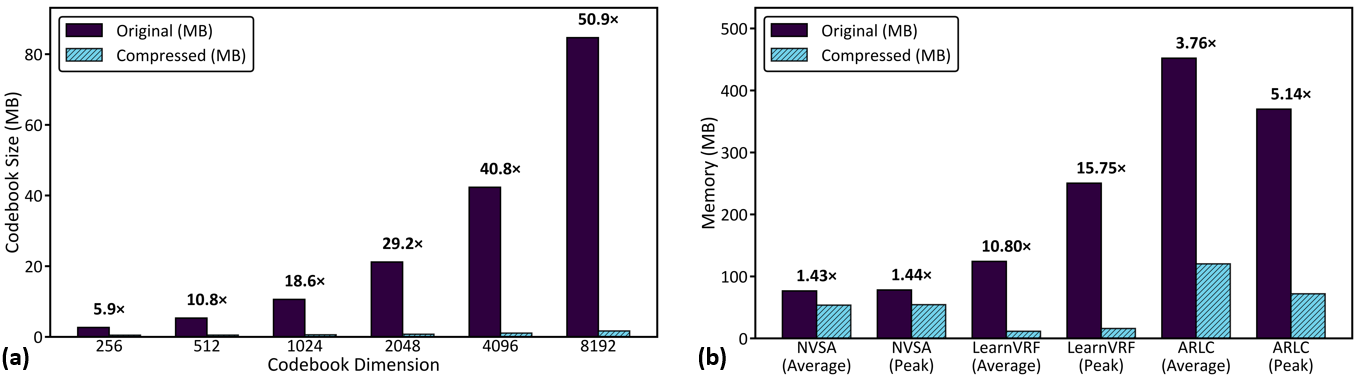}
    \caption{(a) Codebook memory usage and compression ratio across dimensions. (b) Model-level GPU memory usage before and after Gram-Space compression.}
    \label{fig:usage}
    \vspace{-20pt}
\end{figure}

\begin{table}[t!]
\centering
\small
\caption{Memory footprint before and after Gram-Space for top NeSy operators.}
\label{tab:top3_ops}
\begin{tabular}{lr|lr|lr}
\toprule
\multicolumn{2}{c|}{\textbf{Original NVSA}} & \multicolumn{2}{c|}{\textbf{Original LearnVRF}} & \multicolumn{2}{c}{\textbf{Original ARLC}} \\
\cmidrule(lr){1-2} \cmidrule(lr){3-4} \cmidrule(l){5-6}
\textbf{Operator} & \textbf{Mem (KB)} & \textbf{Operator} & \textbf{Mem (KB)} & \textbf{Operator} & \textbf{Mem (KB)} \\
\midrule
Empty $\&$ copy       & 63,929 & Empty $\&$ copy           & 198,359 & Empty $\&$ copy              & 1,290,510 \\
MatMul              & 11,223 & Concat.             & 91,200  & Concat.                & 537,948 \\
To device           & 19,369 & Index   & 66,944  & Circular pad         & 506,880 \\
Element-wise product   & 8,917 & MatMul           & 69,376  & MatMul       & 593,920 \\
\bottomrule
\toprule
\multicolumn{2}{c|}{\textbf{Compressed NVSA}} & \multicolumn{2}{c|}{\textbf{Compressed LearnVRF}} & \multicolumn{2}{c}{\textbf{Compressed ARLC}} \\
\cmidrule(lr){1-2} \cmidrule(lr){3-4} \cmidrule(l){5-6}
\textbf{Operator} & \textbf{Mem (KB)} & \textbf{Operator} & \textbf{Mem (KB)} & \textbf{Operator} & \textbf{Mem (KB)} \\
\midrule
Empty $\&$ copy       & 1,497  & Empty $\&$ copy   & 6,405   & Empty $\&$ copy       & 322,830 \\
MatMul               & 8,430    & Concat.           & 2,844   & Concat.         & 126,720 \\
To device             & 323    & Index             & 2,088   & Circular pad      & 126,720 \\
Element-wise product        & 521    & MatMul   & 10,228   & MatMul       &  156,952 \\
\bottomrule
\end{tabular}%

\end{table}

\vspace{-10pt}
\subsection{Results and Discussion}
To decouple compression benefits from model-specific control flow, we isolate matrix space operators and measure their memory footprints and latency across various original dimensions. As illustrated in Fig.~\ref{fig:usage}(a), Gram-Space successfully compresses VSA codebooks across a wide range of codebook dimensions $D$. This evaluation reports static memory attributed to the codebook and its associated matrix space operators. We observe that the benefit becomes more pronounced as the original vector dimension increases. 
Fig.~\ref{fig:usage}(b) reports model-level memory behavior before and after integrating Gram-Space. We observe reductions in both average and peak allocated GPU memory. These results include dynamic allocations created during codebook-associated stages. The reduction is most pronounced in soft selection pipelines because these pipelines repeatedly execute matrix-based codebook operations and create large intermediate tensors.
As shown in Table~\ref{tab:top3_ops}, the largest memory-consuming operators before compression are dominated by allocation and tensor kernels such as empty and copy memory. This trend indicates that large temporary tensors and layout conversions contribute substantially to memory pressure in codebook-associated stages. After compression, the top operators in each model shift to matrix compute kernels such as matrix multiplication (MatMul), and the reported per-operator memory decreases substantially. This shift is consistent with smaller intermediate tensors in the compressed matrix space representation.

As demonstrated in Table~\ref{tab:codebook_latency}, reducing the dimension of the matrix space representation decreases the overhead of individual similarity computations and improves query latency. 
To provide a clear metric, we measure the computational latency of random attribute lookups in the global codebook across increasing dimensions. Because the query function approximates cosine similarity via matrix multiplication, latency scales almost linearly with the dimension.
As indicated in Table~\ref{tab:latency_speedup}, Gram-Space reduces end-to-end latency across all 3 models, with speedups ranging from $1.68\times$ to $3.62\times$. ARLC shows the largest improvement, consistent with its higher frequency of codebook-associated operations during rule exploration. The decompose overhead varies across models and correlates with how often each model transitions from matrix space to component-wise space. NVSA uses preset rules and triggers only 6 decompose operations per inference pass, while LearnVRF and ARLC require 42 and 90 respectively due to learned and exploratory rule behavior. In all cases, each decompose operation is a small matrix multiplication with per-operation latency under 0.08~ms, and the aggregate overhead remains modest relative to the pipeline savings.

We also compare Gram-Space with existing codebook compression schemes in Table~\ref{tab:comparison_others}. Accuracy after reconstruction is calculated by a similarity score between the original and reconstructed codebook (100\% indicates exact reconstruction). When Gram-Space is applied, the NeSy models after codebook reconstruction do not experience any model accuracy loss on RAVEN and I-RAVEN datasets. Compared to existing techniques applied to the same NeSy codebooks~\cite{sparsehd, FFT-based, dpqhd}, Gram-Space simultaneously achieves superior compression efficiency and accuracy, due to its structure-preserving nature and its ability to effectively exploit the extreme sparsity inherent in VSA high-dimensional vectors.

\begin{table}[t!]
\centering
\small
\caption{Compression effect on similarity query latency across codebook dimensions.}
\label{tab:codebook_latency}
\renewcommand{\arraystretch}{0.95}
\begin{tabular}{|l|c|c|c|c|c|c|}
\hline
 & \textbf{256} & \textbf{512} & \textbf{1024} & \textbf{2048} & \textbf{4096} & \textbf{8192} \\ \hline
\textbf{Original ($\mu$s)}   & 44.7 & 45.3 & 45.7 & 63.8 & 215.9 & 555.6 \\ \hline
\textbf{Compressed ($\mu$s)} & 18.8 & 19.3 & 20.3 & 20.6 & 20.6  & 20.7  \\ \hline
\end{tabular}
\end{table}

\begin{table}[t!]
\centering
\caption{Latency breakdown and speedup analysis.}
\label{tab:latency_speedup}
\small
\begin{tabular}{|l|c|c|c|}
\hline
 & \textbf{NVSA} & \textbf{LearnVRF} & \textbf{ARLC} \\
\hline
Original latency (ms)
  & 44.48 & 43.34 & 190.46 \\
\hline\hline
\multicolumn{4}{|l|}{\textit{With Gram-Space compression}} \\
\hline
\quad Compressed pipeline (ms)
  & 26.08 & 22.33 & 46.01 \\
\hline
\quad Added decompose (ms)
  & 6$\times$0.072 = 0.43
  & 42$\times$0.071 = 2.98
  & 90$\times$0.074 = 6.66 \\
\hline
\quad Compressed total (ms)
  & 26.51 & 25.31 & 52.67 \\
\hline\hline
Speedup
  & \textbf{1.68$\times$}
  & \textbf{1.71$\times$}
  & \textbf{3.62$\times$} \\
\hline
\end{tabular}
\end{table}


\begin{table}[t!]
\centering
\caption{Comparison of similarity preservation and compression performance.}
\label{tab:comparison_others}
\small 
\begin{tabular}{|l|c|c|c|c|}
\hline
~ & \textbf{SparseHD} & \textbf{FFT-based} & \textbf{DPQ-HD} & \textbf{Gram-Space} \\ \hline
\textbf{Similarity score} & 100\% & 10.9\% & 52\% & 100\% \\ \hline
\textbf{Compression ratio} & 1.5$\times$ & 51$\times$ & 6$\times$ & 51$\times$ \\ \hline
\end{tabular}
\end{table}

\begin{figure}[t!]
  \includegraphics[width=\linewidth]{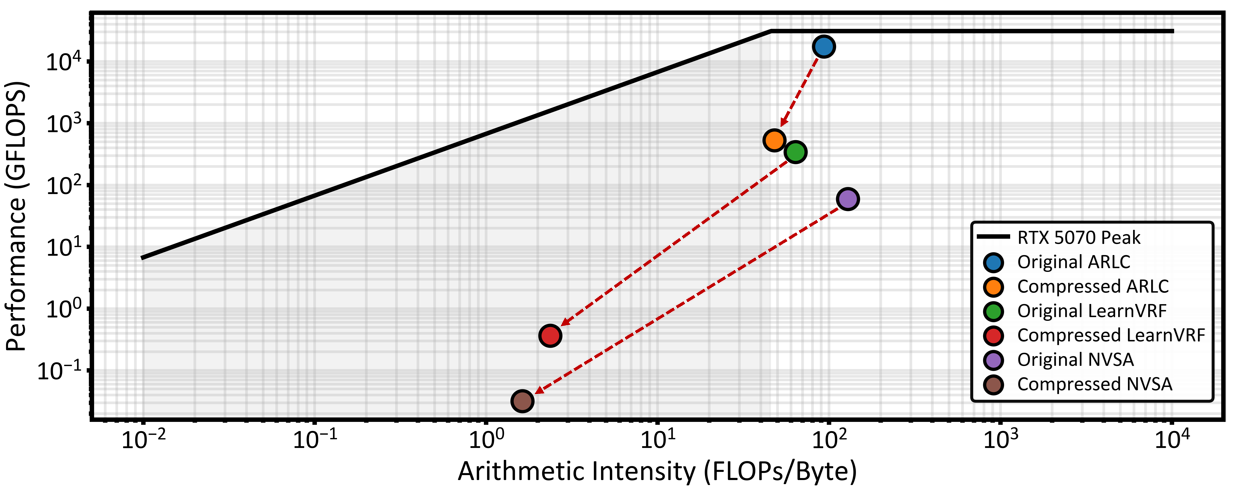}
  \vspace{-10pt}
  \caption{Roofline analysis for NeSy models on RTX 5070.}
  \label{fig:roofline}
  \vspace{-10pt}
\end{figure}

Furthermore, the roofline plot in Fig.~\ref{fig:roofline} shows that all 3 models shift leftward and downward after compression. The downward shift reflects the reduction in total arithmetic work due to smaller matrix dimensions, which is consistent with the latency improvements reported earlier. The leftward shift indicates that the compressed workloads operate at lower arithmetic intensity. For LearnVRF and NVSA, the compressed operating points fall into the memory-bound region of the roofline, which indicates that after codebook compression, the remaining bottleneck is memory bandwidth rather than compute throughput. This suggests that further performance gains require reducing memory traffic in the reconstruction and rule execution stages. From a deployment perspective, the observed memory reductions lower the minimum GPU memory required to execute these workloads, which expands feasibility on consumer-grade GPUs where NeSy reasoning would otherwise be constrained by allocator pressure.

\vspace{-10pt}
\section{Conclusion}

In this work, we addressed memory scalability challenges in VSA-based NeSy models that arise from high-dimensional codebooks and allocation-heavy intermediate tensors. We introduced Gram-Space, a compression framework based on Gram-Schmidt orthogonalization that represents codebook vectors in an orthonormal coordinate system for matrix space operators and restores vectors before component-wise symbolic operators. Our correctness analysis shows that inner-product structure is preserved under orthonormal basis representation, which ensures that matrix space computations yield equivalent results. The invariance property supports assurance arguments that compression does not materially alter matrix space behavior. Empirical evaluations on an NVIDIA RTX 5070 demonstrate that Gram-Space achieves significant memory usage reduction and inference speedups across NVSA, LearnVRF, and ARLC. Hardware profiling further shows that the dominant overhead in codebook-associated stages shifts from allocation and tensor materialization toward matrix computation. These results improve the feasibility of deploying NeSy reasoning workloads on memory-constrained platforms and support further system-level optimization of rule execution and data structures.

\clearpage

\bibliography{./list}

\end{document}